%% file: main.tex
\newif\ifdraft
 \draftfalse

\documentclass[11pt]{article}

\usepackage[letterpaper,margin=1in]{geometry}
\usepackage[parfill]{parskip}

\usepackage{authblk}

\usepackage[toc,page,header]{appendix}

\usepackage[utf8]{inputenc} 
\usepackage[T1]{fontenc}    
\usepackage[colorlinks,citecolor=blue,urlcolor=blue,linkcolor=blue,linktocpage=true]{hyperref}       
\usepackage{url}            
\usepackage{booktabs}       
\usepackage{amsfonts}       
\usepackage{nicefrac}       
\usepackage{microtype}      
\usepackage{xcolor}         
\label{key}
\usepackage{thmtools}

\usepackage{comment}
\usepackage{amsmath}

\usepackage{epsfig}
\usepackage{graphicx}
\usepackage{wrapfig}
\usepackage{subfigure}
\usepackage{multirow}
\usepackage{hyperref}
\usepackage{graphicx}
\usepackage{caption}
\usepackage{array}
\usepackage{bbm}
\usepackage{color}
\usepackage{enumerate}
\usepackage{enumitem}
\setlist{leftmargin=10mm}
\usepackage{mathtools}
\usepackage{amsmath,amssymb,amsthm,bm}
\usepackage{amsfonts,graphicx}
\usepackage{mathrsfs}

\usepackage[ruled,vlined,linesnumbered,noend]{algorithm2e}
\usepackage{algpseudocode}

\usepackage[authoryear]{natbib}

\input{macro}

\input{math_command}

\author[1]{Jiachen T. Wang}
\author[2]{Ruoxi Jia}

\affil[1]{Princeton University}
\affil[2]{Virginia Tech\protect\\
\texttt{\small tianhaowang@princeton.edu}, 
\texttt{\small ruoxijia@vt.edu}
}

\date{}

\title{A Note on ``Efficient Task-Specific Data Valuation for Nearest Neighbor Algorithms''}
\begin{document}



\maketitle

\begin{abstract}
Data valuation is a growing research field that studies the influence of individual data points for machine learning (ML) models. Data Shapley, inspired by cooperative game theory and economics, is an effective method for data valuation. However, it is well-known that the Shapley value (SV) can be computationally expensive. Fortunately, \cite{jia2019efficient} showed that for K-Nearest Neighbors (KNN) models, the computation of Data Shapley is surprisingly simple and efficient. 

In this note, we revisit the work of \cite{jia2019efficient} and propose a more natural and interpretable utility function that better reflects the performance of KNN models. We derive the corresponding calculation procedure for the Data Shapley of KNN classifiers/regressors with the new utility functions. Our new approach, dubbed soft-label KNN-SV, achieves the same time complexity as the original method. 
We further provide an efficient approximation algorithm for soft-label KNN-SV based on locality sensitive hashing (LSH). Our experimental results demonstrate that Soft-label KNN-SV outperforms the original method on most datasets in the task of mislabeled data detection, making it a better baseline for future work on data valuation. 
Our code is available at \url{https://github.com/Jiachen-T-Wang/softlabel-knnsv}. 
\end{abstract}

\section{Introduction}

Data valuation is an emerging research area that aims at measuring how much a given data source contributes to the process of training machine learning (ML) models. In the study of data marketplaces, data valuation is used for ensuring equitable compensation for each data owner. In the study of explainable ML, data valuation serves to identify the training examples that significantly impact certain model behaviors. Inspired by cooperative game theory and economic principles, \cite{ghorbani2019data, jia2019towards} initiated the study of using the Shapley value (SV) as a principled approach for data valuation, which is dubbed as ``Data Shapley''. Since then, many different variants of Data Shapley have been proposed \citep{jia2019efficient, ohrimenko2019collaborative, ghorbani2020distributional, wang2020principled, bian2021energy, kwon2022beta, lin2022measuring, wu2022davinz, karlavs2022data, wang2022data}. The significant number of papers devoted to the study of Data Shapley reflects the effectiveness of this method in appropriately quantifying the contribution of a data point to the training of ML models. The success of Data Shapley can be attributed in part to its unique axiomatic characterizations, which align with the natural fairness requirements of the ML context.

However, the Shapley value is known to be computationally expensive. The number of utility function evaluations required by the exact SV calculation grows exponentially with the number of players, i.e., data points in the ML context. 
While there are many Monte Carlo-based Shapley value estimation algorithms have been proposed \citep{lundberg2017unified, jia2019towards, illes2019estimation, okhrati2021multilinear, burgess2021approximating, mitchell2022sampling, lin2022measuring, wang2023note, kolpaczki2023approximating}, these approaches still require at least thousands of utility function evaluations. 
For ML tasks, evaluating the utility function itself (e.g., computing the validation accuracy of the ML model trained on a given dataset) is already computationally expensive, as it requires training a model from scratch. 
Fortunately, \cite{jia2019efficient} have observed that computing the Data Shapley for K-Nearest Neighbors (KNN), one of the classic yet still popular ML algorithms, is surprisingly easy and efficient. Specifically, \cite{jia2019efficient} show that for unweighted KNN classifiers and regressors, the Shapley value of all $N$ data points can be computed \emph{exactly} in an efficient way, without the complex calculation of a weighted average of all the data points' marginal contributions, as suggested by the original formula introduced in \citet{shapley1953value}. To the best of our knowledge, KNN is the only commonly-used ML model where the exact Data Shapley can be efficiently computed (dubbed as `KNN-SV').

In this note, we revisit the work by \cite{jia2019efficient} and present a refined analysis for the Data Shapley of KNN classifiers and regressors, which we refer to as \emph{soft-label KNN-SV}. Specifically, we propose a more natural and interpretable utility function that better reflects the performance of a soft-label KNN model, compared to the one considered in \cite{jia2019efficient}. We then derive the corresponding calculation procedure for the Data Shapley calculation when the refined utility function is being used. The computation of Soft-label KNN-SV achieves the same time complexity as the one in \cite{jia2019efficient}. 
Furthermore, we present an approximation algorithm for Soft-label KNN-SV based on Locality Sensitivity Hashing (LSH) similar to the one in \cite{jia2019efficient}, which can further improve the computational efficiency. 

Finally, we compare the performance of the newly proposed Soft-label KNN-SV with the original KNN-SV on the task of mislabeled data detection. We demonstrate that Soft-label KNN-SV outperforms the original method on most datasets, indicating that it serves as a better baseline for future work on data valuation. This also highlights the importance of choosing an appropriate utility function in Data Shapley methods.



\section{Background of Data Shapley}
\label{sec:background}

In this section, we formalize the data valuation problem for ML and review the concept of the Shapley value. 

\paragraph{Data Valuation for Machine Learning.} 
We denote a dataset $D := \{ z_i \}_{i=1}^N$ containing $N$ data points. The objective of data valuation is to assign a score to each training data point in a way that reflects their contribution to model training. Denote $\I := \{1, \ldots, N\}$ as the index set. 
To analyze a point's ``contribution'', we define a \emph{utility function} $\U: 2^\I \rightarrow \R$, which maps any subset of the training set to a score indicating the usefulness of the subset. $2^\I$ represents the power set of $\I$, i.e., the collection of all subsets of $\I$, including the empty set and $\I$ itself. 
For classification tasks, a common choice for $\U$ is the validation accuracy of a model trained on the input subset. Formally, for any subset $S \subseteq \I$, we have $\U(S) := \metric(\A( \{ z_i \}_{i \in S} ))$, where $\A$ is a learning algorithm that takes a dataset $\{ z_i \}_{i \in S}$ as input and returns a model. $\metric$ is a metric function that evaluates the performance of a given model, e.g., the accuracy of a model on a hold-out test set. 
The goal of data valuation is to partition $\Utot := \U(\I)$, the utility of the entire dataset, to the individual data point $i \in \I$. 
That is, we want to find a score vector $(\phi_i(\U))_{i \in \I}$ where each $\phi_i(\U)$ represents the payoff for the owner of the data point $i$.

\paragraph{The Shapley Value.} 
The SV \citep{shapley1953value} is a classic concept in cooperative game theory to attribute the total gains generated by the coalition of all players. 
At a high level, it appraises each point based on the (weighted) average utility change caused by adding the point into different subsets. 
Given a utility function $\U(\cdot)$, the Shapley value of a data point $i$ is defined as 
\begin{align}
&\phi_i\left( \U \right) := \frac{1}{N} \sum_{k=1}^{N} {N-1 \choose k-1}^{-1} \sum_{S \subseteq \I \setminus \{i\}, |S|=k-1} \left[ \U(S \cup \{i\}) - \U(S) \right]
\label{eq:shapley-formula}
\end{align}

When the context is clear, we omit the argument and simply write $\phi_i$.  
The popularity of the Shapley value is attributable to the fact that it is the \emph{unique} data value notion satisfying the following four axioms~\citep{shapley1953value}:
\begin{itemize}
    \item Dummy player: if $\U\left(S \cup i\right)=\U(S)+c$ for all $S \subseteq \I \setminus i$ and some $c \in \R$, then $\phi_i\left(\U\right)=c$.
    \item Symmetry: if $\U(S \cup i) = \U(S \cup j)$ for all $S \subseteq \I \setminus \{i, j\}$, then $\phi_i(\U)=\phi_j(\U)$. 
    \item Linearity: For any of two utility functions $\U_1, \U_2$ and any $\alpha_1, \alpha_2 \in \R$, $\phi_i \left( \alpha_{1} \U_{1}+\alpha_{2} \U_{2}\right)=\alpha_{1} \phi_i\left(\U_{1}\right)+$ $\alpha_{2} \phi_i\left( \U_{2}\right)$.
    \item Efficiency: for every $\U$, $\sum_{i \in \I} \phi_i(\U)=\U(\I)$.
\end{itemize}
The difference $\U(S \cup i) - \U(S)$ is often termed the \emph{marginal contribution} of data point $i$ to subset $S \subseteq \I \setminus i$. 
We refer the readers to \citep{ghorbani2019data, jia2019towards} and the references therein for a detailed discussion about the interpretation and necessity of the four axioms in the ML context.

\section{Valuing Data for Soft-label KNN Classification}
\label{sec:soft-label-knn}

We consider the setting of supervised learning. Suppose we are given a set of training data points $D = \{(x_i, y_i)\}_{i=1}^{N}$ and a validation set $D_\test = \{ (x_{\test, i}, y_{\test, i}) \}_{i=1}^{N_\test}$. We also denote $\I = \{1, \ldots, N\}$ the index set for training set $D$. Let $S \subseteq \I$ be a subset of data indices. Denote $\pi^{(S)}(i; x_\test)$ the index (in the full training set $D$) of $i$th closest data point in $S$ to $x_\test$. The distance is measured through some appropriate metric, which we use $\ell_2$ distance throughout this note. 
Thus, $(x_{\pi^{(S)}(1; x_\test)}, x_{\pi^{(S)}(2; x_\test)}, \ldots, x_{\pi^{(S)}(n; x_\test)})$ is a reordering of the training instances in $S$ with respect to their distance from $x_\test$. When the context is clear, we omit the argument $x_\test$ and simply write $\pi^{(S)}(i)$. 


As we mentioned earlier, the utility function $\U(\cdot)$ is usually defined as the validation accuracy of a model trained on the input subset. In \cite{jia2019efficient}, for a given validation set $D_\test$, the utility function of a $K$NN classifier trained on $S$ is defined as 
\begin{align}
    \U(S; D_\test ) := \sum_{(x_\test, y_\test) \in D_\test} \U(S; (x_\test, y_\test) )
    \label{eq:old-util-all}
\end{align}
where we overload the notation slightly and write 
\begin{align}
    \U(S; (x_\test, y_\test) ) := \frac{1}{K} \sum_{j=1}^{\min(K, |S|)} \ind[y_{ \pi^{(S)}(j) } = y_{\test}]
    \label{eq:old-util}
\end{align}

The main result in \cite{jia2019efficient} shows the following:
\begin{theorem}[\cite{jia2019efficient}\footnote{We state a more generalized version which does not require $N \ge K$.}]
\label{thm:from-jia}
Consider the utility function in (\ref{eq:old-util}). Given the test data point $(x_{\test}, y_{\test})$, assume that the input dataset $D = \{(x_i, y_i)\}_{i=1}^N$ is sorted according to $\norm{x_i - x_\test}$ in ascending order. Then, the Shapley value of each training point $\phi_i$ can be calculated recursively as follows: 
\begin{align*}
    \phi_N &= \frac{ \ind[y_N = y_\test] }{ \max(K, N) } \\
    \phi_i &= \phi_{i+1} + \frac{ \ind[y_i = y_\test] - \ind[y_{i+1} = y_\test] }{K} \frac{\min(K, i)}{i}
\end{align*}
\end{theorem}

We can see that the time complexity for computing all of $(\phi_1, \ldots, \phi_N)$ is $O(N \log N)$ (for sorting the training set). After computing the Data Shapley score $\phi_i \left( \U(\cdot; (x_\test, y_\test)) \right)$ for each $(x_\test, y_\test) \in D_\test$, one can compute the Data Shapley corresponding to the utility function on the overall validation set in (\ref{eq:old-util-all}) through the linearity axiom of the Shapley value, i.e., 
\begin{align}
\phi_i \left( \U(\cdot; D_\test) \right) = \sum_{(x_\test, y_\test) \in D_\test} \phi_i \left( \U(\cdot; (x_\test, y_\test)) \right)
\end{align}
In \cite{jia2019efficient}, the utility function in (\ref{eq:old-util}) is justified as the likelihood of a soft-label KNN-classifier in predicting the correct label $y_\test$ for $x_\test$. However, this justification is no longer true when $|S| < K$. 
The actual likelihood in this case is $\frac{1}{|S|} \sum_{j=1}^{|S|} \ind[y_{ \pi^{(S)}(j) } = y_{\test}]$ instead of $ \frac{1}{K} \sum_{j=1}^{|S|} \ind[y_{ \pi^{(S)}(j) } = y_{\test}]$. 
Therefore, in this note, we re-define the utility function here as 
\begin{align}
\U(S; (x_\test, y_\test) ) := 
\begin{cases} 
      \frac{1}{C} & |S|=0 \\
      \frac{1}{\min(K, |S|)} \sum_{j=1}^{\min(K, |S|)} \ind[y_{ \pi^{(S)}(j) } = y_{\test}] & |S|>0 
\end{cases}
\label{eq:new-util-func}
\end{align}
where $C$ is the number of classes for the corresponding classification task. This utility function corresponds to the prediction accuracy of an unweighted, soft-label KNN classifier. When $|S|=0$, we set the utility as the accuracy of random guess. 
Using this utility function, we derive a similar procedure for calculating the SV of KNN classifier where the runtime stays the same as $O(N \log N)$. 
We refer to the Data Shapley when using this utility function as soft-label KNN-SV. 



\begin{theorem}
\label{thm:smooth-classification}
Consider the utility function in (\ref{eq:new-util-func}). Given the test data point $(x_{\test}, y_{\test})$, assume that the input dataset $D = \{(x_i, y_i)\}_{i=1}^N$ is sorted according to $\norm{x_i - x_\test}$ in ascending order. Then, the Shapley value of each training point $\phi_i$ can be calculated recursively as follows: 

\begin{small}
\begin{align*}
    \phi_N  &= \frac{\ind[N \ge 2]}{N} \left( \ind[y_N = y_\test] - \frac{ \sum_{i=1}^{N-1} \ind[y_i = y_\test] }{N-1} \right) \left( \sum_{j=1}^{ \min(K, N) - 1 } \frac{1}{j+1} \right) + \frac{1}{N} \left( \ind[y_N = y_\test] - \frac{1}{C} \right) \\
    \phi_i  &= \phi_{i+1} \\ 
    &~~+ 
    \frac{ \ind[y_i = y_\test] - \ind[y_{i+1} = y_\test] }{N-1} 
    \left[
    \sum_{j=1}^{\min(K, N)} \frac{1}{j} 
    + \frac{\ind[N \ge K]}{K} \left( \frac{ \min(i, K) \cdot (N-1) }{i} - K
    \right)
    \right]
\end{align*}
\end{small}
\end{theorem}

\section{LSH-based Approximation for Soft-label KNN-SV}
\label{sec:lsh}

As we can see from Theorem \ref{thm:from-jia} and \ref{thm:smooth-classification}, for every test data point, the exact computation of the soft-label KNN-SV requires sorting all of the training data points according to their distances to \emph{every} validation data point. 
Hence, similar to the original KNN-SV proposed in \cite{jia2019efficient}, it takes $O(\ntest N\log(N))$ many operations to compute the soft-label KNN-SV for all training data points with respect to the full validation set. If both the training set size $N$ and the test set size $\ntest$ are large, the $N \ntest$ factor in the computational complexity can be prohibitively large. 

To accelerate the computation, \cite{jia2019efficient} propose an efficient approximation algorithm for the original KNN-SV based on the locality-sensitive hashing (LSH) technique. 
In this section, we extend the approximation algorithm to soft-label KNN-SV. 

\subsection{Approximating Soft-label KNN-SV with $K^*$ nearest neighbors only}
Similar to \cite{jia2019efficient}, we identify the following approximation for soft-label KNN-SV when one can only identify $K^*$ nearest neighbors. Hence, instead of sorting the entire training set, one can instead aim for a slightly easier task of identifying the $K^*$ nearest neighbors of the validation data point. 

\begin{theorem}
\label{thm:lsh}
Consider the utility function defined in (\ref{eq:new-util-func}). 
Suppose one can find the $K^*$ nearest neighbors of $x_\test$ where $K^* < N$. When $N \ge \max(2, K)$, the approximation $\widehat \phi$ defined as 
\begin{small}
\begin{align*}
    &\widehat \phi_i = \frac{1}{N} \left(\frac{1}{2} - \frac{1}{C} \right)~~~~~\text{for any}~i \ge K^* \\
    &\widehat \phi_i = \widehat \phi_{i+1} 
    + 
    \frac{ \ind[y_i = y_\test] - \ind[y_{i+1} = y_\test] }{N-1} 
    \left[
    \sum_{j=1}^{\min(K, N)} \frac{1}{j} 
    + \frac{\ind[N \ge K]}{K} \left( \frac{ \min(i, K) \cdot (N-1) }{i} - K
    \right)
    \right]~~~~~\text{for}~i < K^* 
\end{align*}
\end{small}
satisfies 
$
\norm{ \widehat \phi - \phi }_\infty 
\le 
\frac{1}{N} \left( \sum_{j=2}^{K-1} \frac{1}{j+1} \right) + \frac{1}{\max(K^*, K)}
= O \left( \frac{\log K}{N} + \frac{1}{\max(K^*, K)} \right)
$. 
\end{theorem}

Theorem \ref{thm:lsh} indicates that we only need to find $K^*$ many nearest neighbors to obtain an approximation of soft-label KNN-SV with $\ell_\infty$ error $O \left( \frac{\log K}{N} + \frac{1}{\max(K^*, K)} \right)$. 
When $K^*$ is of the order of $\Theta(K)$, the approximation error will be dominated by $\frac{1}{ \max(K^*, K) }$. 


\subsection{Efficiently Finding $K^*$ Nearest Neighbors with LSH}


Efficiently retrieving the nearest neighbors of a query in large-scale databases has been a well-studied problem. Various approximation approaches have been proposed to improve the efficiency of the nearest neighbor search. Among these, Locality Sensitive Hashing (LSH) has been experimentally shown to provide significant speedup for the computation of the original KNN-SV method as reported in \cite{jia2019efficient}.

The LSH algorithm has two hyperparameters: the number of hash tables $L$ and the number of hash bits $M$. 
The algorithm first creates $L$ hash tables. Within each hash table, the algorithm converts each data point $x$ into a set of hash codes using an $M$-bit hash function $h(x) = (h_1(x), \ldots, h_M(x))$. The hash function must satisfy a locality condition, which means that any pair of data points that are close to each other have the same hashed value with high probability, while any pair of data points that are far away from each other have the same hashed values with low probability. A commonly used hash function for LSH is $h(x) = \floor{ \frac{w^T x + b}{r} }$, where $w$ is a vector with entries sampled from $\N(0, 1)$ and $b \sim \unif([0, r])$ \citep{datar2004locality}. 
It has been shown that 
\begin{align*}
\Pr[ h(x) = h(x_\test) ] = f_h( \norm{x - x_\test} )
\end{align*}
where $f_h(y) = \int_0^r \frac{1}{y} f_2(\frac{z}{y}) \left( 1 - \frac{z}{r} \right) dz$, and $f_2$ is the probability density function of the absolute value of standard Gaussian distribution $\N(0, 1)$. 

Algorithm \ref{alg:lsh} outlines the LSH-based approximation algorithm for Soft-label KNN-SV. At a high level, the algorithm preprocesses the dataset by mapping each data point to its corresponding hash values and storing them in hash tables. Then, for every validation data point $(x_\test, y_\test)$, the algorithm computes its hash value and gathers all training data points that collide with it in any of the $L$ hash tables. By an appropriate choice of $L$ and $M$, we can ensure that with high probability, all of the $K^*$ nearest neighbors are among the collided data points. Finally, the algorithm computes an approximation of Soft-label KNN-SV based on Theorem \ref{thm:lsh}.


\begin{algorithm}[t]
\SetAlgoLined
\SetKwInOut{Input}{input}
\SetKwInOut{Output}{output}
\Input{
$L$ -- number of hash tables, 
$M$ -- number of hash bits per table entry. 
}

\CommentSty{// Preprocessing}

Sample a collection of hash functions $\{ h_{\ell, m} \}_{\ell = 1, \ldots, L, m = 1, \ldots, M}$ where each hash function is independently sampled and is of the form $h(x) = \floor{ \frac{w^T x + b}{r} }, w \sim \N(0, I_d), b \sim \unif([0, r])$. 

Initialize $L$ hash tables $\{H_\ell\}_{\ell=1, \ldots, L}$. 

\For{$\ell = 1, \ldots, L$}{
    \For{$i \in \I$}{
        Compute $(h_{\ell, 1}(x_i), \ldots, h_{\ell, M}(x_i))$ and store in hash table $H_\ell$. 
    }
}

\CommentSty{}

\CommentSty{// Find Nearest Neighbors}

\For{ $(x_\test, y_\test) \in D_\test$ }{

$\neighbor \leftarrow \{\}$. 

\For{$\ell = 1, \ldots, L$}{

Compute $(h_{\ell, 1}(x_\test), \ldots, h_{\ell, m}(x_\test))$. 

Add all elements in $H_\ell$ that collide with $x_\test$ to $\neighbor$. 
    
}

}

Remove all repeated elements in $\neighbor$. 

\CommentSty{}

\CommentSty{// Compute Approximate soft-label KNN-SV}

\If{ $|\neighbor| \ge K^*$ }{

Sort elements $x \in \neighbor$ by $\norm{x - x_\test}$. 

Compute the approximated soft-label KNN-SV according to Theorem \ref{thm:lsh} with $K^*$ nearest neighbors found in $\neighbor$. 

}
\Else{
Print ``Fail''.
}

\caption{LSH-based approximation for Soft-label KNN-SV}
\label{alg:lsh}
\end{algorithm}

We now present a theorem that characterizes the success rate of finding the $K^*$ nearest neighbors based on the relationship between the training set and validation data points.\footnote{
\cite{he2012difficulty} introduced a metric known as ``relative contrast'' and linked it to the runtime of LSH in finding the 1-nearest neighbor, which was further extended to the $K$-nearest neighbor by \cite{jia2019efficient}. 
However, the original analysis in \cite{he2012difficulty}'s Theorem 3.1 is vacuous and only applies to imaginary datasets where all training data points have the same distance to every validation point. 
In this note, we present Theorem \ref{thm:refinedlsh}, which provides a refined analysis that corrects the mathematical issues found in \cite{he2012difficulty}.
}

\begin{theorem}
\label{thm:refinedlsh}
For training set $D$, denote $\pi(j; x_\test) := \pi^{(D)}(j; x_\test)$. 
With probability at least $1-\delta$ over the choices of hash functions $\{ h_{\ell, m} \}_{\ell = 1, \ldots, L, m = 1, \ldots, M}$ where each hash function is independently sampled and is of the form $h(x) = \floor{ \frac{w^T x + b}{r} }, w \sim \N(0, I), b \sim \unif([0, r])$, 
Algorithm \ref{alg:lsh} can find \emph{all} of $x_\test$'s $K^*$ nearest neighbors with $M = O\left( \frac{\log N}{\log (1/p_{\max})} \right)$ and $L = O\left( N^c \log(N_\test K^* / \delta) \right)$, where 
    \begin{align*}
    p_1(x_\test) &:= f_h \left(\norm{x_{ \pi(K^*; x_\test) } - x_\test} \right) \\
    p_2(x_\test) &:= f_h \left(\norm{x_{ \pi(K^*+1; x_\test) } - x_\test} \right) \\
    p_{\max} &:= \max_{x_\test} p_2(x_\test ) \\
    c &:= \max_{x_\test} \frac{\log p_1 (x_\test) }{\log p_2 (x_\test)} 
    \end{align*}
    In this setting, there are 
    \begin{align*}
        O(MLN) = O \left(N^{1+c} \log N \frac{\log( N_\test K^*/\delta )}{\log(1/p_{\max})}\right)
    \end{align*}
    hash bits to store, and the expected number of collided data points to check and sort is 
    \begin{align*}
        O\left( N_\test N^c K^* \log( N_\test K^*/\delta) \right) 
    \end{align*}
\end{theorem}

The ratio $c$ in the above theorem determines the space and time complexity of the LSH algorithm. As $f_h$ is monotonically decreasing, we know that $c \leq 1$. Intuitively, $c$ represents the difficulty of finding all $K^*$ nearest neighbors for all data points in a validation set. A smaller $c$ implies that the $(K^*+1)$th nearest neighbor is likely to collide with the validation data, making it more challenging to differentiate between data points that are within $K^*$ nearest neighbors and those that are farther.

It is worth noting that while Algorithm \ref{alg:lsh} reduces the runtime for finding nearest neighbors to each $(x_\test, y_\test)$ to sublinear in $N$, the data preprocessing step increases to $O(MLN) = \widetilde{O}(N^{1+c})$. Therefore, the total runtime becomes $\widetilde{O}(N^{1+c} + N_\test N^c)$. Algorithm \ref{alg:lsh} provides speedups when $N_\test \gg N$.






\section{Extension: Closed-form SV for Soft-label KNN Regression}
\label{sec:knn-sv-regression}

We now extend Theorem \ref{thm:smooth-classification} to unweighted soft-label KNN regression. 
In \cite{jia2019efficient}, the utility function for KNN-regression is  
\begin{align}
    \U(S; (x_\test, y_\test)) = - \left( \frac{1}{K} \sum_{j=1}^{ \min(K, |S|) } y_{\pi^{(S)}(j)} - y_\test \right)^2
\end{align}

Similar to Section \ref{sec:soft-label-knn}, we also consider a more accurate and interpretable utility function for KNN Regression task in the following, and we derive a similar iterative procedure to compute the exact SV for it. 
\begin{align}
\U(S; (x_\test, y_\test)) := 
\begin{cases} 
      - y_\test^2 & |S|=0 \\
      - \left( \frac{1}{\min(K, |S|)} \sum_{j=1}^{\min(K, |S|)} y_{ \pi^{(S)}(j) } - y_{\test} \right)^2 & |S|>0 
\end{cases}
\label{eq:new-util-func-regression}
\end{align}

\begin{theorem}
\label{thm:smooth-regression}
Consider the utility function in (\ref{eq:new-util-func-regression}). Given the test data point $(x_{\test}, y_{\test})$, assume that the input dataset $D = \{(x_i, y_i)\}_{i=1}^N$ is sorted according to $\norm{x_i - x_\test}$ in ascending order. Then, the Shapley value of each training point $\phi_i$ can be calculated recursively as follows: 
\begin{align}
    \phi_i - \phi_{i+1} 
    = \frac{y_{i+1} - y_i}{N-1} \left[ (y_i + y_{i+1}) A_1 + 2 A_2 - 2y_\test A_3 \right]
    \label{eq:correct}
\end{align}
and 
\begin{align}
\phi_N = \frac{1}{N} (*) + \frac{1}{N} \left[ y_\test^2 - (y_N - y_\test)^2 \right]
\end{align}
where 
\begin{align*}
    A_1 &= 
    \left( \sum_{j=1}^K \frac{1}{j^2} \right)
    + \frac{1}{K^2} \left( \frac{(N-1)\min(K, i)}{i} - K \right) \\
    A_2 &= 
    \frac{1}{N-2} \left( \sum_{\ell \in \I \setminus \{i, i+1\} } y_\ell \right) \left( \sum_{j=1}^{K-1} \frac{j}{(j+1)^2} \right) \\
    &~~~+ \frac{1}{K^2} \left( \left( \sum_{\ell=1}^{i-1} y_\ell \right) \left( \frac{(N-1) \min(K, i) \min(K-1, i-1) }{2(i-1)i} - \frac{(K-1)K}{2(N-2)} \right) \right. \\
    &\left.~~~~~~~~~~~~+ \sum_{\ell=i+2}^N y_\ell \left( \frac{(N-1) \min(K, \ell-1) \min(K-1, \ell-2) }{2 (\ell-1)(\ell-2) } - \frac{ (K-1)K }{ 2(N-2) } \right)
    \right) \\
    A_3 &= \left( \sum_{j=1}^K \frac{1}{j} \right) + \min(K, i) \frac{N-1}{i K} - 1
\end{align*}
and 
\begin{align*}
(*) 
&= 
\sum_{j=1}^{K-1}
\left[
\frac{2j+1}{j^2 (j+1)^2} \left( 
\frac{j(j-1)}{(N-1)(N-2)} \left(\sum_{i=1}^{N-1} y_i \right)^2 + \frac{ j(N-j-1) }{ (N-1)(N-2) } \sum_{i=1}^{N-1} y_i^2 
\right) \right. \\
&~~~~~~~~~~~+ 
\left( - \frac{2y_N}{(j+1)^2} - \frac{2 y_\test }{j(j+1)} \right) 
\frac{j}{N-1} \sum_{i=1}^{N-1} y_i \\
&~~~~~~~~~~~+ \left.  \left( \frac{y_N}{ j+1 } - 2y_\test \right) \left( - \frac{y_N}{j+1} \right)  
\right] 
\end{align*}
\end{theorem}



\section{Numerical Evaluation}
\label{sec:experiment}

We compare the effectiveness of Soft-Label KNN-SV and the Original KNN-SV on the task of mislabeled data detection. We use 13 standard datasets that are previously used in the data valuation literature as the benchmark tasks. The description for dataset preprocessing is deferred to Appendix \ref{appendix:settings-dataset}. We generate noisy labeled samples by flipping labels for a randomly chosen 10\% of training data points. We use F1-score as the performance metric for mislabeling detection. 

We consider two different detection rules for mislabeled data: 
\textbf{(1) Ranking} We mark a data point as a mislabeled one if its data value is less than 10 percentile of all data value scores. 
\textbf{(2) Cluster} We use a clustering-based procedure as the number of mislabeled data points and the threshold for detecting noisy samples are usually unknown in practice. Specifically, we first divide all data values into two clusters using the K-Means clustering algorithm and then classify a data point as a noisy sample if its value is less than the minimum of the two cluster centers. This detection rule is adapted from \citet{kwon2022beta}. 

The results are shown in Table \ref{tb:comparison}. We set $K = 5$ for all results (we found that the performance is relatively robust against the choice of $K$). As we can see, Soft-label KNN-SV slightly outperforms the original KNN-SV on most of the datasets. This finding indicates that Soft-label KNN-SV could be considered as a baseline approach for future research on data valuation. 


\begin{table}[t]
\centering
\begin{tabular}{@{}ccccc@{}}
\toprule
\textbf{Dataset}    & \textbf{Original (R)} & \textbf{Soft-Label (R)} & \textbf{Original (C)} & \textbf{Soft-Label (C)} \\ \midrule
\textbf{MNIST}      & 0.86                          & 0.86                            & 0.529                        & 0.529                          \\
\textbf{FMNIST}     & 0.68                          & 0.68                            & 0.571                        & 0.571                          \\
\textbf{CIFAR10}    & 0.18                          & 0.18                            & 0.043                        & 0.043                          \\
\textbf{Fraud}      & 0.775                         & 0.775                           & 0.504                        & 0.504                          \\
\textbf{Creditcard} & 0.23                          & 0.24 \textbf{(+0.1)}                     & 0.23                         & 0.244 \textbf{(+0.014)}                 \\
\textbf{Vehicle}    & 0.375                         & 0.38 \textbf{(+0.05)}                    & 0.197                        & 0.198 \textbf{(+0.001)}                 \\
\textbf{Apsfail}    & 0.675                         & 0.675                           & 0.406                        & 0.412 \textbf{(+0.06)}                  \\
\textbf{Click}      & 0.18                          & 0.19 \textbf{(+0.1)}                     & 0.2                          & 0.206 \textbf{(+0.006)}                 \\
\textbf{Phoneme}    & 0.535                         & 0.545 \textbf{(+0.1)}                    & 0.509                        & 0.516 \textbf{(+0.007)}                 \\
\textbf{Wind}       & 0.475                         & 0.48 \textbf{(+0.005)}                    & 0.416                        & 0.425 \textbf{(+0.009)}                 \\
\textbf{Pol}        & 0.685                         & 0.7 \textbf{(+0.015)}                    & 0.438                        & 0.438                          \\
\textbf{CPU}        & 0.66                          & 0.665 \textbf{(+0.005)}                  & 0.57                         & 0.604 \textbf{(+0.034)}                 \\
\textbf{2DPlanes}   & 0.64                          & 0.64                            & 0.553                        & 0.571 \textbf{(+0.018)}                 \\ \bottomrule
\end{tabular}
\caption{Comparison of mislabel data detection ability of the seven data valuation methods on the 13 classification datasets. \textbf{(R)} denotes the Ranking detection rule, and \textbf{(C)} denotes the Cluster detection rule. }
\label{tb:comparison}
\end{table}

\section{Conclusion}

In this technical note, we present an improved version of KNN-SV which considers a more natural and interpretable utility function for soft-label KNN. We also present a similar LSH-based approximation algorithm for the new KNN-SV, and we provide a refined analysis for the algorithm which eliminates the flawed assumptions made in \cite{he2012difficulty}. 
Moreover, we empirically show that the newly proposed soft-label KNN-SV consistently outperforms the original one. This note advocates using soft-label KNN-SV as a better baseline approach when developing data valuation techniques.

\section*{Acknowledgments}
We thank Yuqing Zhu at UC Santa Barbara for the helpful discussion on the theoretical analysis for ``relative contrast'' in \cite{he2012difficulty}. 
We thank Peter Hannagan Brosten for helping us to identify the typos in Equation (\ref{eq:correct}).

\newpage

\bibliographystyle{plainnat}
\bibliography{ref}


\newpage
\onecolumn

\appendix

\section{Proofs}

\subsection{Proof for Theorem \ref{thm:smooth-classification}}

\begin{customthm}{\ref{thm:smooth-classification}}
Consider the utility function in (\ref{eq:new-util-func}). Given the test data point $(x_{\test}, y_{\test})$, assume that the input dataset $D = \{(x_i, y_i)\}_{i=1}^N$ is sorted according to $\norm{x_i - x_\test}$ in ascending order. Then, the Shapley value of each training point $\phi_i$ can be calculated recursively as follows: 

\begin{small}
\begin{align*}
    \phi_N  &= \frac{\ind[N \ge 2]}{N} \left( \ind[y_N = y_\test] - \frac{ \sum_{i=1}^{N-1} \ind[y_i = y_\test] }{N-1} \right) \left( \sum_{j=1}^{ \min(K, N) - 1 } \frac{1}{j+1} \right) + \frac{1}{N} \left( \ind[y_N = y_\test] - \frac{1}{C} \right) \\
    \phi_i  &= \phi_{i+1} \\ 
    &~~+ 
    \frac{ \ind[y_i = y_\test] - \ind[y_{i+1} = y_\test] }{N-1} 
    \left[
    \sum_{j=1}^{\min(K, N)} \frac{1}{j} 
    + \frac{\ind[N \ge K]}{K} \left( \frac{ \min(i, K) \cdot (N-1) }{i} - K
    \right)
    \right]
\end{align*}
\end{small}
\end{customthm}

\begin{proof}

We first analyze the marginal contribution $\U(S \cup i) - \U(S)$. Without of losing generality, we assume that $D$ is sorted according to $\norm{x_i - x_\test}$. Denote $S_1 = S \cap \{1, \ldots, i-1\}$. 

If $|S| \ge K$, we have 
\begin{align}
    \U(S \cup i) - \U(S) = 
    \begin{cases} 
      0 & |S_1| \ge K \\
      \frac{1}{K} (\ind[y_i = y_{\test}] - \ind[y_{\pi^{(S)}(K)} = y_\test]) & |S_1| < K 
    \end{cases}
\end{align}
If $|S| < K$, we have 
\begin{align}
    \U(S \cup i) - \U(S) = 
    \begin{cases} 
      \frac{1}{|S|+1}\ind[y_i = y_{\test}] 
      + (\frac{1}{|S|+1} - \frac{1}{|S|}) 
      \sum_{j \in S} \ind[y_j = y_\test]
      & |S| > 0 \\
      \ind[y_i = y_{\test}] - \frac{1}{C} & |S| = 0
    \end{cases}
\end{align}

Therefore, for any $S \subseteq I \setminus \{i, i+1\}$, if $|S| \ge K$ and $|S_1| < K$, we have 
\begin{align*}
    \U(S \cup i) - \U(S \cup i+1) 
    &= 
    \frac{1}{K}(\ind[y_i = y_{\test}] - \ind[y_{i+1} = y_{\test}])
\end{align*}
and if $|S| < K$ we have 
\begin{align*}
    \U(S \cup i) - \U(S \cup i+1) 
    &= 
    \begin{cases} 
      \frac{1}{|S|+1}(\ind[y_i = y_{\test}] - \ind[y_{i+1} = y_{\test}])
      & |S| > 0 \\
      \ind[y_i = y_{\test}] - \ind[y_{i+1} = y_{\test}] & |S| = 0
    \end{cases} \\
    &= 
    \frac{1}{|S|+1}(\ind[y_i = y_{\test}] - \ind[y_{i+1} = y_{\test}])
\end{align*}
Overall, for any $S \subseteq I$, we have 
\begin{align}
    \U(S \cup i) - \U(S \cup i+1) 
    = 
    \begin{cases} 
    \frac{1}{\min(|S|+1, K)}(\ind[y_i = y_{\test}] - \ind[y_{i+1} = y_{\test}])
    & |S_1| < K \\
    0 & |S_1| \ge K 
    \end{cases}
\end{align}

Now we analyze the difference between the Shapley value of $i$ and $i+1$. 

\textbf{Case 1: $N > K$.} 
\begin{align*}
    &\phi_i - \phi_{i+1} \\
    &= 
    \frac{1}{N-1} 
    \sum_{j=0}^{N-2} \frac{1}{{N-2 \choose j}} 
    \sum_{S \subseteq \I \setminus \{i, i+1\}, |S|=j} 
    [\U(S \cup i) - \U(S \cup i+1)] \\
    &= 
    \frac{1}{N-1} 
    \left[
    \sum_{j=0}^{K-1} \frac{1}{j+1} (\ind[y_i = y_{\test}] - \ind[y_{i+1} = y_{\test}]) \right. \\
    &~~~~~~~~~~~~+ 
    \left.
    \sum_{j=K}^{N-2} \frac{1}{{N-2 \choose j}} 
    \sum_{ \substack{S_1 \subseteq \{1, \ldots, i-1\},\\ S_2 \subseteq \{i+2, \ldots, N\}, \\ |S_1|+|S_2|=j, |S_1| < K} } 
    \frac{\ind[y_i = y_{\test}] - \ind[y_{i+1} = y_{\test}]}{K}
    \right] \\
    &= \frac{\ind[y_i = y_{\test}] - \ind[y_{i+1} = y_{\test}]}{N-1} 
    \left[
    \left(\sum_{j=1}^{K} \frac{1}{j} \right) + 
    \frac{1}{K} \sum_{j=K}^{N-2} \frac{1}{{N-2 \choose j}} \sum_{m=0}^{K-1} {i-1 \choose m} {N-i-1 \choose j-m}
    \right] 
\end{align*}

Note that 
\begin{align*}
    &\sum_{j=K}^{N-2} \frac{1}{{N-2 \choose j}} \sum_{m=0}^{K-1} {i-1 \choose m} {N-i-1 \choose j-m} \\
    &= \sum_{j=K}^{N-2} \frac{1}{{N-2 \choose j}} \sum_{m=0}^{ \min(j, K-1)} {i-1 \choose m} {N-i-1 \choose j-m} \\
    &= 
    \underbrace{ 
    \sum_{j=0}^{N-2} \frac{1}{{N-2 \choose j}} \sum_{m=0}^{ \min(j, K-1)} {i-1 \choose m} {N-i-1 \choose j-m}}_{(A)}
    - 
    \underbrace{
    \sum_{j=0}^{K-1} \frac{1}{{N-2 \choose j}} \sum_{m=0}^{ \min(j, K-1)} {i-1 \choose m} {N-i-1 \choose j-m} }_{(B)}
\end{align*}

$(A)$ can be simplified as 
\begin{align*}
(A) &= \sum_{m=0}^{\min( K-1, i-1 )} \sum_{j' = 0}^{N-i-1} \frac{ {i-1 \choose m} {N-i-1 \choose j'} }{ {N-2 \choose m+j'} } \\
&= \frac{(N-1)\min(i, K)}{i}
\end{align*}
We can also easily see that $(B) = K$. 

Thus, 
\begin{align*}
    \phi_i - \phi_{i+1} 
    = 
    \frac{\ind[y_i = y_{\test}] - \ind[y_{i+1} = y_{\test}]}{N-1} 
    \left[
    \sum_{j=1}^{K} \frac{1}{j} + 
    \frac{1}{K} \left( \frac{(N-1)\min(i, K)}{i} - K \right)
    \right] 
\end{align*}
and 
\begin{align*}
    \phi_N 
    &= \frac{1}{N} \sum_{j=0}^{K-1} \frac{1}{{N-1 \choose j}} \sum_{S \subseteq \I \setminus \{N\}, |S|=j} [\U(S \cup N) - \U(S)] \\
    &= \frac{1}{N} \left[
    \left(\ind[y_N = y_\test] - \frac{1}{C}\right)
    + 
    \underbrace{
    \sum_{j=1}^{K-1} \frac{1}{{N-1 \choose j}} \sum_{ S \subseteq \I \setminus \{N\}, |S|=j} \left[ 
    \frac{\ind[y_N = y_\test]}{j+1} + \left(\frac{1}{j+1} - \frac{1}{j}\right) 
    \sum_{\ell \in S} \ind[y_\ell = y_\test]
    \right]}_{(*)}
    \right]
\end{align*}

and 
\begin{align*}
    (*) &= 
    \sum_{j=1}^{K-1} \frac{1}{{N-1 \choose j}} \left[
    {N-1 \choose j} \cdot \frac{ \ind[y_N = y_\test] }{j+1}
    - 
    \frac{1}{j(j+1)} \sum_{S \subseteq \I \setminus \{N\}, |S|=j} \sum_{\ell \in S} \ind[y_\ell = y_\test]
    \right] \\
    &= \ind[y_N = y_\test] \sum_{j=1}^{K-1} \frac{1}{j+1} 
    - 
    \underbrace{
    \sum_{j=1}^{K-1} \frac{1}{{N-1 \choose j}} \frac{1}{j(j+1)} \sum_{S \subseteq \I \setminus \{N\}, |S|=j} \sum_{\ell \in S} \ind[y_\ell = y_\test] }_{(**)}
\end{align*}

Denote $s := \sum_{i=1}^{N-1} \ind[y_i = y_\test]$. 
\begin{align*}
\sum_{S \subseteq \I \setminus \{N\}, |S|=j} \sum_{\ell \in S} \ind[y_\ell = y_\test] 
&= 
\sum_{k=0}^{j} k {s \choose k} {N-1-s \choose j-k} \\
&= \sum_{k=1}^{j} k \frac{(s)!}{k! (s-k)!} {N-1-s \choose j-k} \\
&= \sum_{k=1}^{j} \frac{(s)!}{(k-1)! (s-k)!} {N-1-s \choose j-k} \\
&= s \sum_{k=1}^{j} {s-1 \choose k-1} {N-1-s \choose j-k} \\
&= s \sum_{k=0}^{j-1} {s-1 \choose k} {N-1-s \choose j-1-k} \\
&= s \cdot {N-2 \choose j-1}
\end{align*}

Hence, 
\begin{align*}
    (**) 
    &= 
    \sum_{j=1}^{K-1} \frac{1}{{N-1 \choose j}} \frac{1}{j(j+1)} s \cdot {N-2 \choose j-1} \\
    &= s \sum_{j=1}^{K-1} \frac{1}{j(j+1)} \frac{ \frac{(N-2)!}{(j-1)!(N-1-j)!} }{
    \frac{(N-1)!}{j!(N-1-j)!}} \\
    &= \frac{s}{N-1} \sum_{j=1}^{K-1} \frac{1}{j+1} 
\end{align*}
which can be computed in $O(N)$. 
Hence, we have 
\begin{align*}
    \phi_N 
    &= \frac{1}{N} \left[
    \left(\ind[y_N = y_\test] - \frac{1}{C}\right)
    + \ind[y_N = y_\test] \sum_{j=1}^{K-1} \frac{1}{j+1} 
    - \frac{s}{N-1} \sum_{j=1}^{K-1} \frac{1}{j+1} 
    \right] \\
    &= \frac{1}{N} \left[
    \left( \ind[y_N = y_\test] - \frac{s}{N-1}  \right) \sum_{j=1}^{K-1} \frac{1}{j+1} 
    +
    \left(\ind[y_N = y_\test] - \frac{1}{C}\right)
    \right]
\end{align*}

\textbf{Case 2: $1 \le N \le K$.}

\begin{align*}
    \phi_i - \phi_{i+1} 
    &= 
    \frac{1}{N-1} 
    \sum_{j=0}^{N-2} \frac{1}{{N-2 \choose j}} 
    \sum_{S \subseteq \I \setminus \{i, i+1\}, |S|=j} 
    [\U(S \cup i) - \U(S \cup i+1)] \\
    &= 
    \frac{1}{N-1} 
    \left[
    \sum_{j=0}^{N-2} \frac{1}{j+1} (\ind[y_i = y_{\test}] - \ind[y_{i+1} = y_{\test}]) \right] \\
    &= \frac{\ind[y_i = y_{\test}] - \ind[y_{i+1} = y_{\test}]}{N-1} 
    \sum_{j=1}^{N-1} \frac{1}{j}
\end{align*}
and 
\begin{align*}
    \phi_N
    &= \frac{1}{N} \sum_{j=0}^{N-1} \frac{1}{{N-1 \choose j}} \sum_{S \subseteq \I \setminus \{N\}, |S|=j } [\U(S \cup N) - \U(S)] \\
    &= \frac{1}{N} \sum_{j=1}^{N-1} \frac{1}{{N-1 \choose j}} \sum_{S \subseteq \I \setminus \{N\}, |S|=j } \left[ \frac{\ind[y_N = y_\test]}{j+1} + \left( \frac{1}{j+1} - \frac{1}{j} \right) \sum_{\ell \in S}\ind[y_\ell = y_\test] \right] \\
    &~~~~+ \frac{1}{N} \left( \ind[y_N = y_\test] - \frac{1}{C} \right) \\
    &= \frac{1}{N} \sum_{j=1}^{N-1} \frac{1}{{N-1 \choose j}} \left[ {N-1 \choose j} \frac{\ind[y_N = y_\test]}{j+1} - \frac{1}{(j+1)j} \sum_{S \subseteq \I \setminus \{N\}, |S|=j }
    \sum_{\ell \in S}\ind[y_\ell = y_\test] \right] \\
    &= \frac{\ind[y_N = y_\test]}{N} \sum_{j=1}^{N-1} \frac{1}{j+1} 
    - 
    \frac{1}{N} \sum_{j=1}^{N-1} \frac{1}{ {N-1 \choose j} } \frac{1}{j(j+1)} s {N-2 \choose j-1} 
    + \frac{1}{N} \left( \ind[y_N = y_\test] - \frac{1}{C} \right) \\
    &= \frac{1}{N} \left( \ind[y_N = y_\test] - \frac{s}{N-1} \right) \sum_{j=1}^{N-1} \frac{1}{j+1} 
    + \frac{1}{N} \left( \ind[y_N = y_\test] - \frac{1}{C} \right) 
\end{align*}
\end{proof}

\newpage

\subsection{Proof for Theorem \ref{thm:lsh}}

\begin{customthm}{\ref{thm:lsh}}
Consider the utility function defined in (\ref{eq:new-util-func}). 
Suppose one can find the $K^*$ nearest neighbors of $x_\test$ where $K^* < N$. When $N \ge \max(2, K)$, the approximation $\widehat \phi$ defined as 
\begin{small}
\begin{align*}
    &\widehat \phi_i = \frac{1}{N} \left(\frac{1}{2} - \frac{1}{C} \right)~~~~~\text{for any}~i \ge K^* \\
    &\widehat \phi_i = \widehat \phi_{i+1} 
    + 
    \frac{ \ind[y_i = y_\test] - \ind[y_{i+1} = y_\test] }{N-1} 
    \left[
    \sum_{j=1}^{\min(K, N)} \frac{1}{j} 
    + \frac{\ind[N \ge K]}{K} \left( \frac{ \min(i, K) \cdot (N-1) }{i} - K
    \right)
    \right]~~~~~\text{for}~i < K^* 
\end{align*}
\end{small}
satisfies 
$
\norm{ \widehat \phi - \phi }_\infty 
\le 
\frac{1}{N} \left( \sum_{j=2}^{K-1} \frac{1}{j+1} \right) + \frac{1}{\max(K^*, K)}
= O \left( \frac{\log K}{N} + \frac{1}{\max(K^*, K)} \right)
$. 
\end{customthm}

\begin{proof}


Denote $s := \sum_{i=1}^{N-1} \ind[y_i = y_\test]$. 
Since $N \ge K$, we have 
\begin{align*}
\phi_N &= \frac{1}{N} \left(\ind[y_N = y_\test] - \frac{s}{N-1}\right) \left( \sum_{j=1}^{K-1} \frac{1}{j+1} \right) + \frac{1}{N} \left( \ind[y_N = y_\test] - \frac{1}{C} \right) \\ 
\phi_i &= \phi_{i+1} + \frac{ \ind[y_i = y_\test] - \ind[y_{i+1} = y_\test] }{N-1} \left( \sum_{j=1}^{K-1} \frac{1}{j+1} + \frac{ \min(i, K) (N-1) }{iK} \right)
\end{align*}

Since $0 \le s \le N-1$, we have 
\begin{align*}
- \frac{1}{N} \left( \sum_{j=1}^{K-1} \frac{1}{j+1} \right) - \frac{1}{NC} \le 
\phi_N
\le \frac{1}{N} \left( \sum_{j=1}^{K-1} \frac{1}{j+1} \right) + \frac{1}{N}\left(1-\frac{1}{C}\right)
\end{align*}

For any $i \ge K$, we have 
\begin{align*}
\phi_i 
&= \phi_N + \sum_{\ell=i}^{N-1} (\phi_\ell - \phi_{\ell+1}) \\
&= \frac{1}{N} \left(\ind[y_N = y_\test] - \frac{s}{N-1}\right) \left( \sum_{j=1}^{K-1} \frac{1}{j+1} \right) + \frac{1}{N} \left( \ind[y_N = y_\test] - \frac{1}{C} \right) \\
&~~~+ \sum_{\ell=i}^{N-1} 
\frac{ \ind[y_\ell = y_\test] - \ind[y_{\ell+1} = y_\test] }{N-1} 
\left( \sum_{j=1}^{K-1} \frac{1}{j+1} + \frac{ N-1 }{\ell} \right) \\
&= \frac{1}{N} \left(\ind[y_N = y_\test] - \frac{ \sum_{\ell=1}^{N-1} \ind[y_\ell = y_\test] }{N-1}\right) \left( \sum_{j=1}^{K-1} \frac{1}{j+1} \right) + \frac{1}{N} \left( \ind[y_N = y_\test] - \frac{1}{C} \right) \\
&~~~+ 
\frac{ \ind[y_i = y_\test] - \ind[y_{N} = y_\test] }{N-1} \left( \sum_{j=1}^{K-1} \frac{1}{j+1} \right) + 
\sum_{\ell=i}^{N-1} 
\frac{ \ind[y_\ell = y_\test] - \ind[y_{\ell+1} = y_\test] }{ \ell } \\
&= \frac{1}{N} \left(\ind[y_N = y_\test] - \frac{ \sum_{\ell=1}^{N-1} \ind[y_\ell = y_\test] }{N-1}\right) \left( \sum_{j=1}^{K-1} \frac{1}{j+1} \right) - \frac{1}{CN} \\
&~~~+ 
\frac{ \ind[y_i = y_\test] - \ind[y_{N} = y_\test] }{N-1} \left( \sum_{j=1}^{K-1} \frac{1}{j+1} \right) \\
&~~~+ \frac{\ind[y_i = y_\test]}{i} - \sum_{\ell=i}^{N-1} \frac{1}{\ell(\ell+1)} \ind[y_{\ell+1} = y_\test] \\
&= \left( \sum_{j=1}^{K-1} \frac{1}{j+1} \right) \left[ 
\frac{\ind[y_i = y_\test]}{N-1} - \frac{ \sum_{\ell=1}^N \ind[y_\ell = y_\test] }{ N(N-1) }
\right] \\
&~~~+ \frac{\ind[y_i = y_\test]}{i} - \sum_{\ell=i}^{N-1} \frac{1}{\ell(\ell+1)} \ind[y_{\ell+1} = y_\test] - \frac{1}{CN} 
\end{align*}

By inspection, we know that $\phi_i$ is being maximized when $\ind[y_i = y_\test] = 1$ and $\ind[y_\ell = y_\test] = 0$ for all $\ell \ne i$, hence 
\begin{align*}
\phi_i \le \frac{1}{N} \left( \sum_{j=1}^{K-1} \frac{1}{j+1} \right) + \frac{1}{i} - \frac{1}{CN}
\end{align*}
and $\phi_i$ is being minimized when $\ind[y_i = y_\test] = 0$ and $\ind[y_\ell = y_\test] = 1$ for all $\ell \ne i$, hence 
\begin{align*}
\phi_i \ge 
- \frac{1}{N} \left( \sum_{j=1}^{K-1} \frac{1}{j+1} \right) - \left( \frac{1}{i} - \frac{1}{N} \right) - \frac{1}{CN}
\end{align*}

For any $i < K$, we have 
\begin{align*}
\phi_i 
&= \phi_K + \sum_{\ell=i}^{K-1} (\phi_\ell - \phi_{\ell+1}) \\
&= \phi_K + 
\sum_{\ell=i}^{K-1} \left( \frac{ \ind[y_i = y_\test] - \ind[y_{i+1} = y_\test] }{N-1} \left( \sum_{j=1}^{K-1} \frac{1}{j+1} + \frac{ N-1 }{K} \right) \right) \\
&= \phi_K + \left( \sum_{j=1}^{K-1} \frac{1}{j+1} + \frac{ N-1 }{K} \right) \frac{ \ind[y_i = y_\test] - \ind[y_K = y_\test] }{N-1} \\
&= \left( \sum_{j=1}^{K-1} \frac{1}{j+1} \right) \left[ 
\frac{\ind[y_K = y_\test]}{N-1} - \frac{ \sum_{\ell=1}^N \ind[y_\ell = y_\test] }{ N(N-1) }
\right] \\
&~~~+ \frac{\ind[y_K = y_\test]}{K} - \sum_{\ell=K}^{N-1} \frac{1}{\ell(\ell+1)} \ind[y_{\ell+1} = y_\test] - \frac{1}{CN} \\
&~~~+ \left( \sum_{j=1}^{K-1} \frac{1}{j+1} + \frac{ N-1 }{K} \right) \frac{ \ind[y_i = y_\test] - \ind[y_K = y_\test] }{N-1} \\
&= \left( \sum_{j=1}^{K-1} \frac{1}{j+1} \right) \left[ 
\frac{\ind[y_i = y_\test]}{N-1} - \frac{ \sum_{\ell=1}^N \ind[y_\ell = y_\test] }{ N(N-1) }
\right] \\
&~~~+ \frac{\ind[y_i = y_\test]}{K} - \sum_{\ell=K}^{N-1} \frac{1}{\ell(\ell+1)} \ind[y_{\ell+1} = y_\test] - \frac{1}{CN} 
\end{align*}

By inspection, we know that $\phi_i$ is being maximized when $\ind[y_i = y_\test] = 1$ and $\ind[y_\ell = y_\test] = 0$ for all $\ell \ne i$, hence 
\begin{align*}
\phi_i \le \frac{1}{N} \left( \sum_{j=1}^{K-1} \frac{1}{j+1} \right) + \frac{1}{K} - \frac{1}{CN}
\end{align*}
and $\phi_i$ is being minimized when $\ind[y_i = y_\test] = 0$ and $\ind[y_\ell = y_\test] = 1$ for all $\ell \ne i$, hence 
\begin{align*}
\phi_i \ge 
- \frac{1}{N} \left( \sum_{j=1}^{K-1} \frac{1}{j+1} \right) - \left( \frac{1}{K} - \frac{1}{N} \right) - \frac{1}{CN}
\end{align*}

Hence, by setting $\widehat \phi_i = \frac{1}{N}\left(\frac{1}{2} - \frac{1}{C}\right)$, for all $i \ge K^*$, we will have 
\begin{align*}
| \phi_i - \widehat \phi_{i} |
&\le 
\frac{1}{N} \left( \sum_{j=1}^{K-1} \frac{1}{j+1} \right) + \frac{1}{\max(i, K)} - \frac{1}{CN} - \left( \frac{1}{N}\left(\frac{1}{2} - \frac{1}{C}\right) \right) \\
&= \frac{1}{N} \left( \sum_{j=1}^{K-1} \frac{1}{j+1} \right) + \frac{1}{\max(i, K)} - \frac{1}{2N}
\end{align*}

For any $i < K^*$, since $\phi_i - \phi_{i+1} = \widehat \phi_i - \widehat \phi_{i+1}$, we have 
$\phi_i - \widehat \phi_i = \phi_{i+1} - \widehat \phi_{i+1}$, hence we will have 
\begin{align*}
|\phi_i - \widehat \phi_i| 
\le \frac{1}{N} \left( \sum_{j=1}^{K-1} \frac{1}{j+1} \right) + \frac{1}{\max(K^*, K)} - \frac{1}{2N}
\end{align*}

Hence we have 
\begin{align*}
    \norm{ \phi_i - \widehat \phi_i }_\infty 
    &\le 
    \frac{1}{N} \left( \sum_{j=1}^{K-1} \frac{1}{j+1} \right) + \frac{1}{\max(K^*, K)} - \frac{1}{2N} \\
\end{align*}

\end{proof}

\newpage

\subsection{Proof for Theorem \ref{thm:refinedlsh}}

\begin{customthm}{\ref{thm:refinedlsh}}
For training set $D$, denote $\pi(j; x_\test) := \pi^{(D)}(j; x_\test)$. 
With probability at least $1-\delta$ over the choices of hash functions $\{ h_{\ell, m} \}_{\ell = 1, \ldots, L, m = 1, \ldots, M}$ where each hash function is independently sampled and is of the form $h(x) = \floor{ \frac{w^T x + b}{r} }, w \sim \N(0, I), b \sim \unif([0, r])$, 
Algorithm \ref{alg:lsh} can find \emph{all} of $x_\test$'s \emph{all} $K^*$ nearest neighbors with $M = O\left( \frac{\log N}{\log (1/p_{\max})} \right)$ and $L = O\left( N^c \log(N_\test K^* / \delta) \right)$, where 
    \begin{align*}
    p_1(x_\test) &:= f_h \left(\norm{x_{ \pi(K^*; x_\test) } - x_\test} \right) \\
    p_2(x_\test) &:= f_h \left(\norm{x_{ \pi(K^*+1; x_\test) } - x_\test} \right) \\
    p_{\max} &:= \max_{x_\test} p_2(x_\test ) \\
    c &:= \max_{x_\test} \frac{\log p_1 (x_\test) }{\log p_2 (x_\test)} 
    \end{align*}
    In this setting, there are 
    \begin{align*}
        O(MLN) = O \left(N^{1+c} \log N \frac{\log( N_\test K^*/\delta )}{\log(1/p_{\max})}\right)
    \end{align*}
    hash bits to store, and the expected number of collided data points to check and sort is 
    \begin{align*}
        O\left( N_\test N^c K^* \log( N_\test K^*/\delta) \right) 
    \end{align*}
\end{customthm}

\begin{proof}

For any given $x_\test$, we want to analyze the probability of finding its $K^*$ nearest neighbors among $\I$. 
For notation convenience, we assume that the input dataset $D = \{(x_i, y_i)\}_{i=1}^N$ is sorted according to $\norm{x_i - x_\test}$ in ascending order, and we omit the argument and denote 
\begin{align*}
    p_1 &:= \Pr_h \left[ h(x_{K^*}) = h(x_\test) \right] = f_h( \norm{x_{K^*} - x_\test} ), \\ 
    p_2 &:= \Pr_h \left[ h(x_{K^*+1}) = h(x_\test) \right] = f_h( \norm{x_{K^*+1} - x_\test} )
\end{align*}

For any $1 \le i \le K^*$, we have 
\begin{align*}
    \Pr[ x_i \text{ not in any hash table} ] \le (1 - p_1^m)^L
\end{align*}
and for any $i \ge K^*+1$, we have 
\begin{align*}
    \Pr[ x_i \text{ in a hash table} ] \le p_2^m
\end{align*}
where the probability is taken over all hash functions. 

We want the probability where there exists $1 \le i \le K^*$ such that $x_i \notin \neighbor$ is small. That is, we want 
\begin{align*}
    K^* (1 - p_1^m)^L \le \delta
\end{align*}
which leads to 
\begin{align}
    L \ge \log( \delta / K^* ) / \log(1 - p_1^m) 
\end{align}

The expected number of data points in $\neighbor$, in this case, can be upper-bounded as 
\begin{align}
    \E\left[ | \neighbor | \right] \le K^* L + (N-K^*) L p_2^m
\end{align}
and we would like to choose $m$ such that the expected runtime is sublinear in $N$. One way to do that is by setting $(N-K^*) p_2^m \le N p_2^m = O(1)$, which makes $p_2^m = O(1/N)$. 
Denote $c := \frac{\log p_1}{\log p_2}$. 
We have 
\begin{align*}
    p_1^m = (p_2^m)^c = O(1/N^c)
\end{align*}
Hence, we have 
\begin{align*}
    \log( \delta / K^* ) / \log(1 - p_1^m) 
    \le \frac{\log(K^* / \delta)}{ p_1^m }
    = O\left( N^c \log(K^*/\delta) \right)
\end{align*}
Hence, when we set 
\begin{align*}
    L &:= O\left( N^c \log(K^*/\delta) \right) \\
    m &:= \frac{ \log(N) }{ \log(1/p_2) }
\end{align*}
, the expected number of data points to check is upper bounded by
\begin{align*}
(K^* + O(1)) L 
= O\left( N^c K^* \log(K^*/\delta) \right) 
\end{align*}

Now, we want the above property to hold for all test data points with probability at least $1 - \delta$, which we simply set $\delta \leftarrow \delta / N_\test$ and take the maximum in terms of all possible $x_\test$ for quantities in the above analysis that depend on $x_\test$.  
\end{proof}

\newpage

\subsection{Proof for Theorem \ref{thm:smooth-regression}}

\begin{customthm}{\ref{thm:smooth-regression}}
Consider the utility function in (\ref{eq:new-util-func-regression}). Given the test data point $(x_{\test}, y_{\test})$, assume that the input dataset $D = \{(x_i, y_i)\}_{i=1}^N$ is sorted according to $\norm{x_i - x_\test}$ in ascending order. Then, the Shapley value of each training point $\phi_i$ can be calculated recursively as follows: 
\begin{align}
    \phi_i - \phi_{i+1} 
    = \frac{y_{i+1} - y_i}{N-1} \left[ (y_i + y_{i+1}) A_1 + 2 A_2 - 2y_\test A_3 \right]
\end{align}
where 
\begin{align*}
    A_1 &= \sum_{j=1}^K \frac{1}{j^2} + \frac{1}{K^2} \left( \frac{(N-1)\min(K, i)}{i} - K \right) \\
    A_2 &= 
    \frac{1}{N-2} \left( \sum_{\ell \in \I \setminus \{i, i+1\} }^N y_\ell \right) \left( \sum_{j=1}^{K-1} \frac{j}{(j+1)^2} \right) \\
    &~~~+ \frac{1}{K^2} \left( \left( \sum_{\ell=1}^{i-1} y_\ell \right) \left( \frac{(N-1) \min(K, i) \min(K-1, i-1) }{2(i-1)i} - \frac{(K-1)K}{2(N-2)} \right) \right. \\
    &\left.~~~~~~~~~~~~+ \sum_{\ell=i+2}^N y_\ell \left( \frac{(N-1) \min(K, \ell-1) \min(K-1, \ell-2) }{2 (\ell-1)(\ell-2) } - \frac{ (K-1)K }{ 2(N-2) } \right)
    \right) \\
    A_3 &= \left( \sum_{j=1}^K \frac{1}{j} \right) + \min(K, i) \frac{N-1}{i K} - 1
\end{align*}
and 
\begin{align*}
(*) 
&= 
\sum_{j=1}^{K-1}
\left[
\frac{2j+1}{j^2 (j+1)^2} \left( 
\frac{j(j-1)}{(N-1)(N-2)} \left(\sum_{i=1}^{N-1} y_i \right)^2 + \frac{ j(N-j-1) }{ (N-1)(N-2) } \sum_{i=1}^{N-1} y_i^2 
\right) \right. \\
&~~~~~~~~~~~+ 
\left( - \frac{2y_N}{(j+1)^2} - \frac{2 y_\test }{j(j+1)} \right) 
\frac{j}{N-1} \sum_{i=1}^{N-1} y_i \\
&~~~~~~~~~~~+ \left.  \left( \frac{y_N}{ j+1 } - 2y_\test \right) \left( - \frac{y_N}{j+1} \right)  
\right] 
\end{align*}
\end{customthm}
\begin{proof}
We first derive the marginal contribution of a data point $i$ to the base dataset $S$. Let $S_1 = S \cap \{1, \ldots, i-1\}$. 

If $|S_1| \ge K$, we have $v(S \cup i) - v(S) = 0$, which implies that $v(S \cup i) - v(S \cup i+1) = 0$. 

If $|S_1| < K$, we have
\begin{align*}
&v(S \cup i) - v(S \cup i+1) \\
&= \left( \frac{1}{ \min(|S|+1, K) } \sum_{\ell=1}^{\min(|S|+1, K)} y_{\pi^{(S \cup i+1)}(\ell)} - y_\test \right)^2 - \left( \frac{1}{ \min(|S|+1, K) } \sum_{\ell=1}^{\min(|S|+1, K)} y_{\pi^{(S \cup i)}(\ell)} - y_\test \right)^2 \\
&= 
\frac{ y_{i+1} - y_i }{ \min(|S|+1, K) } 
\left(
\frac{ y_i + y_{i+1} + 2 \sum_{\ell=1}^{\min(|S|, K-1)} y_{\pi^{(S)}( \ell )} }{ \min(|S|+1, K) } - 2y_\test
\right)
\end{align*}

Now we can analyze $\phi_i - \phi_{i+1}$. 
\begin{align*}
&\phi_i - \phi_{i+1} \\
&= \frac{1}{N-1} \sum_{j=0}^{N-2} \frac{1}{{N-2 \choose j}} \sum_{S \subseteq \I \setminus \{i, i+1\}, |S|=j} [\U(S \cup i) - \U(S \cup i+1)] \\
&= 
\frac{1}{N-1} \sum_{j=0}^{N-2} \frac{1}{{N-2 \choose j}} \sum_{ \substack{S_1 \subseteq \{1, \ldots, i-1\},\\ S_2 \subseteq \{i+2, \ldots, N\}, \\ |S_1|+|S_2|=j, \\ |S_1| \le K-1} } \left[
\frac{ y_{i+1} - y_i }{ \min(j+1, K) } 
\left(
\frac{ y_i + y_{i+1} + 2 \sum_{\ell=1}^{\min(j, K-1)} y_{\pi^{(S)}(\ell)} }{ \min(j+1, K) } - 2y_\test
\right)
\right] \\
&= 
\frac{y_{i+1} - y_i}{N-1} \sum_{j=0}^{N-2} \frac{1}{{N-2 \choose j}} \sum_{ \substack{S_1 \subseteq \{1, \ldots, i-1\},\\ S_2 \subseteq \{i+2, \ldots, N\}, \\ |S_1|+|S_2|=j, \\ |S_1| \le K-1} } \left[
\frac{ y_i + y_{i+1} + 2 \sum_{\ell=1}^{ \min(j, K-1) } y_{\pi^{(S)}(\ell)} }{ \min(j+1, K)^2 } - \frac{ 2y_\test }{ \min(j+1, K) }
\right] \\
&= 
\frac{y_{i+1} - y_i}{N-1} \left[
(y_i + y_{i+1}) 
\underbrace{
\sum_{j=0}^{N-2} \frac{ 1 }{{N-2 \choose j} \cdot \min(j+1, K)^2 } \sum_{ \substack{S_1 \subseteq \{1, \ldots, i-1\},\\ S_2 \subseteq \{i+2, \ldots, N\}, \\ |S_1|+|S_2|=j, \\ |S_1| \le K-1} } (1) }_{(*)} \right. \\
&~~~~~~~~~~~~~~~~~~~
+ 2 
\underbrace{
\sum_{j=0}^{N-2} \frac{ 1 }{{N-2 \choose j} \cdot \min(j+1, K)^2 } \sum_{ \substack{S_1 \subseteq \{1, \ldots, i-1\},\\ S_2 \subseteq \{i+2, \ldots, N\}, \\ |S_1|+|S_2|=j, \\ |S_1| \le K-1} } \sum_{\ell=1}^{ \min(j, K-1) } y_{\pi^{(S)}(\ell)} }_{(**)} \\
&~~~~~~~~~~~~~~~~~~~\left.
- 2 y_\test 
\underbrace{
\sum_{j=0}^{N-2} \frac{ 1 }{{N-2 \choose j} \cdot \min(j+1, K) } \sum_{ \substack{S_1 \subseteq \{1, \ldots, i-1\},\\ S_2 \subseteq \{i+2, \ldots, N\}, \\ |S_1|+|S_2|=j, \\ |S_1| \le K-1} } (1)}_{(***)}
\right]
\end{align*}

A simple calculation shows that 
\begin{align*}
\sum_{ \substack{S_1 \subseteq \{1, \ldots, i-1\},\\ S_2 \subseteq \{i+2, \ldots, N\}, \\ |S_1|+|S_2|=j, \\ |S_1| \le K-1} } (1) = 
\begin{cases} 
      {N-2 \choose j} & 1 \le j \le K-1 \\
      \sum_{m=0}^{K-1} {i-1 \choose m} { N-i-1 \choose j-m } & j \ge K
\end{cases}
\end{align*}

\begin{align*}
(*)
& = 
\sum_{j=0}^{K-1} \frac{ 1 }{ {N-2 \choose j} \cdot (j+1)^2 } {N-2 \choose j} 
+ \sum_{j=K}^{N-2} \frac{ 1 }{ {N-2 \choose j} \cdot K^2 }
\sum_{m=0}^{K-1} {i-1 \choose m} { N-i-1 \choose j-m } \\
&= 
\sum_{j=0}^{K-1} \frac{ 1 }{ (j+1)^2 } 
+ \frac{1}{K^2} 
\sum_{j=K}^{N-2} \frac{ 1 }{ {N-2 \choose j} }
\sum_{m=0}^{K-1} {i-1 \choose m} { N-i-1 \choose j-m } \\
&= 
\left( \sum_{j=1}^{K} \frac{ 1 }{ j^2 } \right)
+ \frac{1}{K^2} \left( \min(K, i) \cdot \frac{N-1}{i} - K \right)
\end{align*}

where we use the identity $\sum_{j=K}^{N-2} \frac{ 1 }{ {N-2 \choose j} } \sum_{m=0}^{K-1} {i-1 \choose m} { N-i-1 \choose j-m } = \min(K, i) \cdot \frac{N-1}{i} - K$ proved in the previous theorem. 

\begin{align*}
& 
(**) \\
&= 
\underbrace{
\sum_{j=0}^{K-1} \frac{ 1 }{{N-2 \choose j} \cdot (j+1)^2 } \sum_{ \substack{S_1 \subseteq \{1, \ldots, i-1\},\\ S_2 \subseteq \{i+2, \ldots, N\}, \\ |S_1|+|S_2|=j } } \sum_{\ell=1}^{ j } y_{\pi^{(S)}(\ell)} }_{(A)}
+ 
\frac{1}{K^2} 
\underbrace{
\sum_{j=K}^{N-2} \frac{ 1 }{{N-2 \choose j} } \sum_{ \substack{S_1 \subseteq \{1, \ldots, i-1\},\\ S_2 \subseteq \{i+2, \ldots, N\}, \\ |S_1|+|S_2|=j, \\ |S_1| \le K-1} } \sum_{\ell=1}^{ K-1 } y_{\pi^{(S)}(\ell)} }_{(B)} 
\end{align*}

For any fixed $j$, denote
\begin{align*}
\sum_{ \substack{S_1 \subseteq \{1, \ldots, i-1\},\\ S_2 \subseteq \{i+2, \ldots, N\}, \\ |S_1|+|S_2|=j, \\ |S_1| \le K-1} } \sum_{\ell=1}^{ \min(j, K-1) } y_{\pi^{(S)}(\ell)}
= 
\sum_{\ell=1}^{i-1} c_\ell y_\ell + \sum_{\ell=i+2}^{N} c_\ell y_\ell 
\end{align*}
where $c_\ell$ is the count of $\ell$th data point that is being added in the summation. By a simple calculation, we know that 
\begin{align*}
c_\ell 
= 
\begin{cases} 
\sum_{m=0}^{\min(j-1, K-2)} { i-2 \choose m } { N-i-1 \choose j-1-m } &~~~ 1 \le \ell \le i-1 \\
\sum_{m=0}^{\min(j-1, K-2)} {\ell-3 \choose m} { N-\ell \choose j-1-m } &~~~ i+2 \le \ell \le N
\end{cases}
\end{align*}

Hence, 
\begin{align*}
(A) 
&= \sum_{j=0}^{K-1} \frac{ 1 }{{N-2 \choose j} \cdot (j+1)^2 } {N-3 \choose j-1} 
\sum_{\ell \in \I \setminus \{i, i+1\}} y_\ell \\
&= \left( \sum_{\ell \in \I \setminus \{i, i+1\}} y_\ell \right) \sum_{j=0}^{K-1} \frac{ {N-3 \choose j-1} }{{N-2 \choose j} \cdot (j+1)^2 } \\
&= \left( \sum_{\ell \in \I \setminus \{i, i+1\}} y_\ell \right) \sum_{j=0}^{K-1} \frac{ j }{ (j+1)^2 (N-2) } \\
&= 
\frac{ \sum_{\ell \in \I \setminus \{i, i+1\}} y_\ell }{N-2} \sum_{j=0}^{K-1} \frac{ j }{ (j+1)^2 }
\end{align*}

\begin{align*}
(B) 
&= 
\sum_{j=K}^{N-2} \frac{ 1 }{{N-2 \choose j} } 
\left[
\left(\sum_{\ell = 1}^{i-1} y_\ell \right) \sum_{m=0}^{K-2} { i-2 \choose m } { N-i-1 \choose j-1-m } 
+ \sum_{\ell = i+2}^{N} y_\ell \sum_{m=0}^{K-2} {\ell-3 \choose m} { N-\ell \choose j-1-m }
\right] \\
&= \left(\sum_{\ell = 1}^{i-1} y_\ell \right) 
\underbrace{
\sum_{j=K}^{N-2} \frac{ 1 }{{N-2 \choose j} } \sum_{m=0}^{K-2} { i-2 \choose m } { N-i-1 \choose j-1-m }}_{(B.1)}
+ 
\underbrace{
\sum_{j=K}^{N-2} \frac{ 1 }{{N-2 \choose j} } 
\sum_{\ell = i+2}^{N} y_\ell \sum_{m=0}^{K-2} {\ell-3 \choose m} { N-\ell \choose j-1-m } }_{(B.2)}
\end{align*}

\begin{align*}
\sum_{j=0}^{N-2} \frac{ 1 }{{N-2 \choose j} } \sum_{m=0}^{K-2} { i-2 \choose m } { N-i-1 \choose j-1-m }
&= \sum_{m=0}^{K-2} \sum_{j=0}^{N-2} 
\frac{ { i-2 \choose m } { N-i-1 \choose j-1-m } }{ {N-2 \choose j} } \\
&= \sum_{m=0}^{K-2} \sum_{j=m+1}^{N-2} 
\frac{ { i-2 \choose m } { N-i-1 \choose j-1-m } }{ {N-2 \choose j} } \\
&= \sum_{m=0}^{K-2} \sum_{j=0}^{N-2 - (m+1)} 
\frac{ { i-2 \choose m } { N-i-1 \choose j } }{ {N-2 \choose j + m+1} }  \\
&= \sum_{m=0}^{ \min(i-2, K-2) } \sum_{j=0}^{N-2 - (m+1)} 
\frac{ { i-2 \choose m } { N-i-1 \choose j } }{ {N-2 \choose j + m+1} } \\
&= \sum_{m=0}^{ \min(i-2, K-2) } \sum_{j=0}^{ N-i-1 } 
\frac{ { i-2 \choose m } { N-i-1 \choose j } }{ {N-2 \choose j + m+1} } \\
&= \sum_{m=0}^{ \min(i-2, K-2) } \frac{ (m+1) (N-1) }{ (i-1) i } \\
&= \frac{N-1}{(i-1)i} \frac{ \min(i, K) \min(i-1, K-1) }{2}
\end{align*}

\begin{align*}
(B.1) 
&= 
\sum_{j=0}^{N-2} \frac{ 1 }{{N-2 \choose j} } \sum_{m=0}^{K-2} { i-2 \choose m } { N-i-1 \choose j-1-m }
- \sum_{j=0}^{K-1} \frac{ 1 }{{N-2 \choose j} } \sum_{m=0}^{K-2} { i-2 \choose m } { N-i-1 \choose j-1-m } \\
&= \frac{N-1}{(i-1)i} \frac{ \min(i, K) \min(i-1, K-1) }{2} - 
\sum_{j=0}^{K-1} \frac{ {N-3 \choose j-1} }{ {N-2 \choose j} } \\
&= \frac{N-1}{(i-1)i} \frac{ \min(i, K) \min(i-1, K-1) }{2} - \sum_{j=0}^{K-1} \frac{j}{N-2} \\
&= \frac{N-1}{(i-1)i} \frac{ \min(i, K) \min(i-1, K-1) }{2} - \frac{ K(K-1) }{2(N-2)} 
\end{align*}

\begin{align*}
(B.2) 
&= \sum_{\ell = i+2}^{N} y_\ell 
\sum_{j=K}^{N-2} \frac{ 1 }{{N-2 \choose j} } 
 \sum_{m=0}^{K-2} {\ell-3 \choose m} { N-\ell \choose j-1-m } \\
&= \sum_{\ell = i+2}^{N} y_\ell 
\left[
\frac{N-1}{(\ell-1)(\ell-2)} \frac{ \min(\ell-1, K) \min(\ell-2, K-1) }{2} - \frac{ K(K-1) }{2(N-2)} 
\right]
\end{align*}

Hence, 
\begin{align*}
(**) 
&= \frac{ \sum_{\ell \in \I \setminus \{i, i+1\}} y_\ell }{N-2} \sum_{j=0}^{K-1} \frac{ j }{ (j+1)^2 } \\
&~~~+ 
\frac{1}{K^2} 
\left(\sum_{\ell = 1}^{i-1} y_\ell \right) \left( 
\frac{N-1}{(i-1)i} \frac{ \min(i, K) \min(i-1, K-1) }{2} - \frac{ K(K-1) }{2(N-2)} \right) \\ 
&~~~+ 
\frac{1}{K^2} 
\sum_{\ell = i+2}^{N} y_\ell 
\left[
\frac{N-1}{(\ell-1)(\ell-2)} \frac{ \min(\ell-1, K) \min(\ell-2, K-1) }{2} - \frac{ K(K-1) }{2(N-2)} 
\right]
\end{align*}

\begin{align*}
(***) 
&= 
\sum_{j=0}^{N-2} \frac{ 1 }{{N-2 \choose j} \cdot \min(j+1, K) } \sum_{ \substack{S_1 \subseteq \{1, \ldots, i-1\},\\ S_2 \subseteq \{i+2, \ldots, N\}, \\ |S_1|+|S_2|=j, \\ |S_1| \le K-1} } (1) \\
&= 
\sum_{j=0}^{K-1} \frac{ 1 }{j+1} + \frac{1}{K} \left( \min(K, i) \frac{N-1}{i} - K \right)
\end{align*}

Overall, we have 
\begin{align*}
\phi_i - \phi_{i+1}
&= \frac{y_{i+1} - y_i}{N-1} \left[
(y_i + y_{i+1}) (*)
+ 2 (**)
- 2y_\test (***)
\right]
\end{align*}

Now we analyze $\phi_N$:

\begin{align*}
\phi_N 
&= \frac{1}{N} \sum_{j=0}^{K-1} \frac{1}{ {N-1 \choose j} } \sum_{ |S|=j, S \subseteq \I \setminus \{N\} } [ v(S \cup N) - v(S) ] \\
&= \frac{1}{N} 
\underbrace{
\sum_{j=1}^{K-1} \frac{1}{ {N-1 \choose j} } \sum_{ |S|=j, S \subseteq \I \setminus \{N\} } \left[
\left( \frac{1}{j} \sum_{i \in S} y_i - y_\test \right)^2 - 
\left( \frac{1}{j+1} \sum_{ i \in S \cup \{N\} } y_i - y_\test \right)^2
\right] 
}_{(*)} \\
&~~~ + 
\frac{1}{N} \left[ y_\test^2 - (y_N - y_\test)^2 \right] 
\end{align*}

\begin{align*}
(*) 
&= 
\sum_{j=1}^{K-1} \frac{1}{ {N-1 \choose j} } \sum_{ |S|=j, S \subseteq \I \setminus \{N\} } \left[
\left( \frac{1}{j} \sum_{i \in S} y_i + \frac{1}{j+1} \sum_{ i \in S \cup \{N\} } y_i - 2 y_\test \right) 
\left( \frac{1}{j} \sum_{i \in S} y_i - \frac{1}{j+1} \sum_{ i \in S \cup \{N\} } y_i \right)
\right] \\
&= 
\sum_{j=1}^{K-1} \frac{1}{ {N-1 \choose j} } \sum_{ |S|=j, S \subseteq \I \setminus \{N\} } \left[
\left( \left(\frac{1}{j} + \frac{1}{j+1}\right) \sum_{i \in S} y_i + \frac{y_N}{j+1} - 2 y_\test \right) 
\left( \frac{1}{j(j+1)} \sum_{i \in S} y_i - \frac{y_N}{j+1} \right)
\right] \\
&= 
\sum_{j=1}^{K-1} \frac{1}{ {N-1 \choose j} } \sum_{ |S|=j, S \subseteq \I \setminus \{N\} } \left[
\left( \frac{2j+1}{j(j+1)}  \sum_{i \in S} y_i + \frac{y_N}{j+1} - 2 y_\test \right) 
\left( \frac{1}{j(j+1)} \sum_{i \in S} y_i - \frac{y_N}{j+1} \right)
\right] \\
&= 
\sum_{j=1}^{K-1} \frac{1}{ {N-1 \choose j} } \sum_{ |S|=j, S \subseteq \I \setminus \{N\} } \left[
\frac{2j+1}{j^2 (j+1)^2} \left( \sum_{i \in S} y_i \right)^2 \right. \\
&~~~~~~~~~~~~~~~~~~~~~~~~~~~~~~~~~~~~+ 
\left( - \frac{2j+1}{j (j+1)^2} y_N + \frac{1}{j(j+1)} \left( \frac{y_N}{j+1} - 2y_\test \right) \right)\left( \sum_{i \in S} y_i \right) \\
&~~~~~~~~~~~~~~~~~~~~~~~~~~~~~~~~~~~~+ \left.
\left( \frac{y_N}{ j+1 } - 2y_\test \right) \left( - \frac{y_N}{j+1} \right) \right] \\
&= 
\sum_{j=1}^{K-1} \frac{1}{ {N-1 \choose j} } 
\left[
\frac{2j+1}{j^2 (j+1)^2}  \sum_{ |S|=j, S \subseteq \I \setminus \{N\} } \left( \sum_{i \in S} y_i \right)^2 \right. \\
&~~~~~~~~~~~~~~~~~~~~~+ 
\left( - \frac{2j+1}{j (j+1)^2} y_N + \frac{1}{j(j+1)} \left( \frac{y_N}{j+1} - 2y_\test \right) \right) \sum_{ |S|=j, S \subseteq \I \setminus \{N\} } \left( \sum_{i \in S} y_i \right) \\
&~~~~~~~~~~~~~~~~~~~~~+ \left.  \left( \frac{y_N}{ j+1 } - 2y_\test \right) \left( - \frac{y_N}{j+1} \right) \sum_{ |S|=j, S \subseteq \I \setminus \{N\} } (1)
\right] 
\end{align*}

\begin{align*}
\sum_{ |S|=j, S \subseteq \I \setminus \{N\} } \left( \sum_{i \in S} y_i \right) 
= {N-2 \choose j-1} \sum_{i=1}^{N-1} y_i
\end{align*}

\begin{align*}
\sum_{ |S|=j, S \subseteq \I \setminus \{N\} } \left( \sum_{i \in S} y_i \right)^2 
&= 
\sum_{ |S|=j, S \subseteq \I \setminus \{N\} } \left( \sum_{i \in S} y_i^2 + \sum_{i, \ell \in S, i \ne \ell} y_i y_\ell \right) \\
&= 
{N-2 \choose j-1} \sum_{i=1}^{N-1} y_i^2 
+ {N-3 \choose j-2} \sum_{i, \ell \in S, i \ne \ell} y_i y_\ell 
\end{align*}

Hence, 
\begin{align*}
(*)
&= 
\sum_{j=1}^{K-1} \frac{1}{ {N-1 \choose j} } 
\left[
\frac{2j+1}{j^2 (j+1)^2} \left( {N-2 \choose j-1} \sum_{i=1}^{N-1} y_i^2 
+ {N-3 \choose j-2} \sum_{i, \ell \in \I \setminus \{N\}, i \ne \ell} y_i y_\ell \right) \right. \\
&~~~~~~~~~~~~~~~~~~~~~+ 
\left( - \frac{2j+1}{j (j+1)^2} y_N + \frac{1}{j(j+1)} \left( \frac{y_N}{j+1} - 2y_\test \right) \right) 
{N-2 \choose j-1} \sum_{i=1}^{N-1} y_i \\
&~~~~~~~~~~~~~~~~~~~~~+ \left.  \left( \frac{y_N}{ j+1 } - 2y_\test \right) \left( - \frac{y_N}{j+1} \right) {N-1 \choose j} 
\right] \\
&= 
\sum_{j=1}^{K-1}
\left[
\frac{2j+1}{j^2 (j+1)^2} \left( \frac{j}{N-1} \sum_{i=1}^{N-1} y_i^2 
+ \frac{j(j-1)}{(N-1)(N-2)} \sum_{i, \ell \in \I \setminus \{N\}, i \ne \ell} y_i y_\ell \right) \right. \\
&~~~~~~~~~~~+ 
\left( - \frac{2j+1}{j (j+1)^2} y_N + \frac{1}{j(j+1)} \left( \frac{y_N}{j+1} - 2y_\test \right) \right) 
\frac{j}{N-1} \sum_{i=1}^{N-1} y_i \\
&~~~~~~~~~~~+ \left.  \left( \frac{y_N}{ j+1 } - 2y_\test \right) \left( - \frac{y_N}{j+1} \right)  
\right] \\
&= 
\sum_{j=1}^{K-1}
\left[
\frac{2j+1}{j^2 (j+1)^2} \left( \frac{j}{N-1} \sum_{i=1}^{N-1} y_i^2 
+ \frac{j(j-1)}{(N-1)(N-2)}  \left( \left(\sum_{i=1}^{N-1} y_i \right)^2 - \sum_{i=1}^{N-1} y_i^2 \right)  \right) \right. \\
&~~~~~~~~~~~+ 
\left( - \frac{2j+1}{j (j+1)^2} y_N + \frac{1}{j(j+1)} \left( \frac{y_N}{j+1} - 2y_\test \right) \right) 
\frac{j}{N-1} \sum_{i=1}^{N-1} y_i \\
&~~~~~~~~~~~+ \left.  \left( \frac{y_N}{ j+1 } - 2y_\test \right) \left( - \frac{y_N}{j+1} \right)  
\right] \\
&= 
\sum_{j=1}^{K-1}
\left[
\frac{2j+1}{j^2 (j+1)^2} \left( 
\frac{j(j-1)}{(N-1)(N-2)} \left(\sum_{i=1}^{N-1} y_i \right)^2 + \frac{ j(N-j-1) }{ (N-1)(N-2) } \sum_{i=1}^{N-1} y_i^2 
\right) \right. \\
&~~~~~~~~~~~+ 
\left( - \frac{2y_N}{(j+1)^2} - \frac{2 y_\test }{j(j+1)} \right) 
\frac{j}{N-1} \sum_{i=1}^{N-1} y_i \\
&~~~~~~~~~~~+ \left.  \left( \frac{y_N}{ j+1 } - 2y_\test \right) \left( - \frac{y_N}{j+1} \right)  
\right] 
\end{align*}

Hence, 
\begin{align*}
\phi_N = \frac{1}{N} (*) + \frac{1}{N} \left[ y_\test^2 - (y_N - y_\test)^2 \right]
\end{align*}
\end{proof}

\newpage

\section{Datasets used in Section \ref{sec:experiment}}
\label{appendix:settings-dataset}
A comprehensive list of datasets and sources is summarized in Table \ref{tb:datasets}. For Fraud, Creditcard, Vehicle, and all datasets from OpenML, we subsample the dataset to balance positive and negative labels. For the image dataset CIFAR10, we follow the common procedure in prior works \citep{ghorbani2019data, jia2019towards, kwon2022beta}: we extract the penultimate layer outputs from the pre-trained ResNet18 \citep{he2016deep}. The pre-training is done with the ImageNet dataset \citep{deng2009imagenet} and the weight is publicly available from PyTorch. We choose features from the class of Dog and Cat. The extracted outputs have dimension 512.

\begin{table}[h]
\centering
\begin{tabular}{@{}cc@{}}
\toprule
\textbf{Dataset} & \textbf{Source}                        \\ \midrule
MNIST            & \citet{lecun1998mnist}                \\
FMNIST           & \citet{xiao2017fashion}               \\
CIFAR10          & \citet{krizhevsky2009learning}        \\
Click            & \url{https://www.openml.org/d/1218}  \\
Fraud            & \citet{dal2015calibrating}            \\
Creditcard       & \citet{yeh2009comparisons}            \\
Vehicle          & \citet{duarte2004vehicle}             \\
Apsfail          & \url{https://www.openml.org/d/41138} \\
Phoneme          & \url{https://www.openml.org/d/1489}  \\
Wind             & \url{https://www.openml.org/d/847}   \\
Pol              & \url{https://www.openml.org/d/722}   \\
CPU              & \url{https://www.openml.org/d/761}   \\
2DPlanes         & \url{https://www.openml.org/d/727}   
\\ \bottomrule
\end{tabular}
\caption{A summary of datasets used in Section \ref{sec:experiment}'s experiments.}
\label{tb:datasets}
\end{table}

\end{document}

%% file: macro.tex
\newif\iffinal

\iffinal
    \newcommand{\tianhao}[1]{}
    \newcommand{\ruoxi}[1]{}
\else
    \newcommand{\tianhao}[1]{{\bf \textcolor{purple}{[Tianhao: #1]}}}
    \newcommand{\ruoxi}[1]{{\bf \textcolor{purple}{[Ruoxi:#1]}}}
\fi

%% file: math_command.tex
\usepackage{cleveref}
\usepackage{bbm}
\newtheorem{theorem}{Theorem}

\newtheorem{remark-star}{Remark}
\newtheorem{remark-star-1}{Remark}

\newtheorem*{proof-sketch}{Proof Sketch}
\usepackage{booktabs}

\newtheorem{innercustomthm}{Theorem}
\newenvironment{customthm}[1]
  {\renewcommand\theinnercustomthm{#1}\innercustomthm}
  {\endinnercustomthm}

\newcommand{\E}{\mathbb{E}}

\newcommand{\R}{\mathbb{R}}

\newcommand{\norm}[1]{\left\lVert#1\right\rVert}
\newcommand{\N}{\mathcal{N}}
\newcommand{\I}{\mathcal{I}}

\newcommand{\ind}{\mathbbm{1}}

\newcommand{\U}{v}

\newcommand{\test}{\mathrm{test}}

\usepackage{tikz}

\def\floor#1{\lfloor #1 \rfloor}

\newcommand{\unif}{\mathrm{Unif}}

\newcommand{\ntest}{N_{\mathrm{test}}}

\newcommand{\neighbor}{\texttt{neighbor}}

\newcommand{\metric}{\texttt{acc}}

\newcommand{\A}{\mathcal{A}}

\newcommand{\Utot}{U_{\mathrm{tot}}}